\documentclass{article}

\usepackage{arxiv}

\usepackage[utf8]{inputenc} 
\usepackage[T1]{fontenc}    
\usepackage{hyperref}       
\usepackage{url}            
\usepackage{booktabs}       
\usepackage{amsfonts}       
\usepackage{nicefrac}       
\usepackage{microtype}      
\usepackage{lipsum}
\usepackage{graphicx}
\graphicspath{ {./images/} }

\title{TTDAsweep: A Novel Dimensionality Reduction Method for Image Classification Tasks}

\author{
 Yu-Shih Chen \\
  Department of Computer Science\\
  University of California, Davis\\
  Davis, CA 95616 \\
   \And
 Melissa Goh \\
  Oracle Corporation\\
  Seattle, WA 98101 \\
  \And
 Norm Matloff (corresponding author)\\
  Department of Computer Science\\
  University of California, Davis\\
  Davis, CA 95616 \\
}

\begin{document}
\maketitle

\section{Summary}
One of the most celebrated achievements of modern machine learning technology is automatic classification of images. However, success is typically achieved only with major computational costs. Here we introduce \textbf{TDAsweep}, a machine learning tool aimed at improving the efficiency of automatic classification of images.

\section{Motivation}
Machine learning technology has become ubiquitous, in business, engineering, the sciences, medicine and so on. One of the most visible types of applications is image classification. Much research is being conducted for instance in medical applications, such as detection of cancer from histology slide data.

In principle, image classification should not be different from any other classification application. Any of the usual methods, say logistic regression, support vector machines or neural networks might be used, taking as features the pixel values. But in practice, the situation quickly spirals out of control as the image size grows. Even with the famous MNIST image test bed \cite{lecun2010mnist}, consisting of very small 28 x 28 images, such an approach implies that the number of features is $28^2$ = 784. Hence an urgent need for \emph{dimension reduction}, i.e. reducing the number of features.

A popular form of dimension reduction in imaging settings is use of \emph{convolutional} methods, which adapt classical image tiling techniques \cite{chollet_allaire_2018}. With proper tuning, these can work well, but at a cost of extensive computation time and large memory footprint. Thus an alternative approach to dimensional reduction is desirable. Here we introduce a new method, TDAsweep, along with software implementing it.

\section{Topological Data Analysis}
\textbf{TDAsweep} is a variant of \emph{topological data analysis} (TDA) \cite{doi:10.1146/annurev-statistics-031017-100045}, a feature extraction technique. In the case of image analysis, TDA forms a \emph{filtration}, a family of geometric overlays of the image. typically involves pixel intensity thresholding, as follows:

Consider \emph{r x s} images in which pixel intensity varies from 0 to 255 (for simplicity, assumed here to be grayscale). Set a threshold \emph{t}, say 100. At each pixel whose intensity is below \emph{t}, set the intensity to 0. The remaining nonzero pixels then will clump into one or more connected sets, in which two pixels are in the same set if they are adjacent horizontally, vertically or diagonally. As we vary \emph{t}, connected sets can grow, shrink, coalesce and subdivide, with the set counts growing and shrinking correspondingly. These counts, \emph{Betti numbers}, form our new, dimension-reduced feature set \cite{garside_henderson_makarenko_masoller_2019}.

\section{TDAsweep}
Our \textbf{TDAsweep} method is inspired both by TDA and by \emph{Run Length Run Number} (RLRN) methods \cite{inproceedings} \cite{mir_hanmandlu_tandon_1995}. In RLRN, the counts are of the numbers of consecutive pixels above a threshold, within a given direction.

TDAsweep should be viewed as a blend between standard TDA and RLRN. For a given threshold t and a given one of the six directions, we count the number of connected components. These counts form the new features.

TDAsweep casts thresholding on the original image (each pixel value above the threshold will be denoted as 1, 0 otherwise). Then, TDAsweep counts contiguous components in horizontal, vertical, and the two diagonal directions of a pixel matrix. The counts of the components in each direction will serve as the new set of features describing the original image.

\subsection{Toy Example}
An example should help illustrate the process more clearly: After thresholding some toy image, say we have the following matrix:
\begin{verbatim}
                        10011101
                        10111100
                        10101101
\end{verbatim}

Then, we would count the number of components in each row, column, and diagonal. An example of counting the components for each row would be:
\begin{verbatim}
        There are 3 components in vector 10011101
        There are 2 components in vector 10111100
        There are 4 components in vector 10101101
\end{verbatim}

Here, [3,2,4] will be included as the new set of features. We repeat this process for columns and the two diagonal directions (NW to SE and NE to SW), for each \emph{t}.

\section{Goals of the Method and Software}
\textbf{TDAsweep} is aimed at reducing computation time while retaining classification accuracy. By first performing dimension reduction using \textbf{TDAsweep}, it is hoped to attain a major reduction in the subsequent model fit time.

\subsection*{Tuning Parameters}
Compared to the "C" part of CNN, \textbf{TDAsweep} has rather few tuning parameters/hyperparameters:

\begin{itemize}
    \item the threshold values (number and levels)
    \item intervalWidth: a coalescing factor
    
\end{itemize}
The latter parameter controls further dimension reduction beyond the initial sweep operation. Say we have an \emph{r x s} image, with this parameter set to \emph{w}. Partition the rows into \emph{r/w} groups of consecutive rows. Then within each group, replace the \emph{w} component counts by just their sum. Do the same for columns and diagonals.

\subsection{Parallelization Feature}
TDAsweep provides the user with the option to parallelize the process. To enable the feature, simply specify the desired number of cores, via the arument \emph{cls}. If parallelization is not needed, set \emph{cls=NULL}. Comparisons of run-time difference between parallel and non-parallel TDAsweep is shown in the showcases below.

\section{Example: MNIST Data}

\emph{MNIST} is an image dataset consisting of 28x28 grayscale handwritten digit images; It is one of the most widely known image datasets for classification.

As one of the case studies, we compared the results of using only the Support Vector Machine (SVM) without any form of dimensionality reduction and the results of using \emph{TDAsweep} before SVM for classification tasks on \emph{MNIST} (refer to Table \ref{mnist}). We used the standard \textbf{e1071} package in R with the default hyperparameters.

\begin{table}[!ht]
\centering
\caption{{\bf SVM + TDAsweep on MNIST.} A comparison between the performance of the SVM model with and without TDAsweep as dimensionality reduction and feature extraction.\label{mnist}}
\begin{tabular}{|l|l|l|l|l|l|l|}
\hline
\multicolumn{1}{|l|}{} & \multicolumn{1}{|l|}{\bf MNIST (raw)} & \multicolumn{1}{|l|}{\bf MNIST (sweep)} & \multicolumn{1}{|l|}{\bf MNIST (sweep)}\\ \hline
No. of features & 784 & 84 & 168  \\ \hline
intervalWidth & NA & 2 & 2  \\ \hline
thresholds & NA & (100) & (100,175)  \\ \hline
TDAsweep total time & NA & 42 min & 78 min  \\ \hline
TDAsweep (core=4) & NA & 12.4 min & 21.7 min  \\ \hline
SVM train time & 123 min & 14 min & 24 min  \\ \hline
Accuracy (Validation Set) & 97.9\% & 97\% & 97.86\% \\ \hline
\end{tabular}
\label{mnist}
\end{table}

As we can see from Table \ref{mnist}, with one threshold and \textbf{intervalWidth} equal to 2, we were able to achieve 89.3\% feature reduction (784 features to 84 features) in exchange for less than 1\% accuracy loss. With some tuning, specifically with 2 thresholds, we achieved a 78.6\% feature reduction in exchange for a mere 0.04\% accuracy loss. \textbf{TDAsweep} drastically decreased the SVM training time, from 123 minutes to 14 and 24 minutes for 1 and 2 thresholds, with a corresponding reduction in overall time from 123 minutes to 56 minutes and 102 minutes, respectively.

\section{Experiments \& Characteristic Analysis}
The above timings were for the \textbf{e1071} R package implementation of SVM. This is the "standard" implementation, but a lesser-known package, \textbf{liquidSVM}, can be much faster. The experiments described below were conducted with this version of SVM. We have used the default parameters for the \textbf{liquidSVM} package as well.

Other than \emph{MNIST}, we analyzed the multifaceted advantages that \emph{TDAsweep} could bring with other widely known image datasets such as \emph{histology-MNIST} \cite{kather_weis_bianconi_melchers_schad_gaiser_marx_zöllner_2016} and \emph{CIFAR-10} \cite{Krizhevsky09learningmultiple}.

\section{TDAsweep Versus Image Augmentation}
Image augmentation is a technique used to generate new image sample data from the existing ones, through random geometric transformations such as horizontal and vertical flips, rotations, scale transformations and so on. They are of particular interest in applications in which the images have no inherent up/down, left/right orientation, such as those from histology slides. The reasoning is that a new image to be classified may be similar to a rotated version, say, of one of the images in the training set, even though not similar to any of those images themselves.

Data augmentation in image processing became well-known after the development of AlexNet \cite{10.5555/2999134.2999257}. Numerous augmented files were added, multiplying the number of images by a factor of 2048. The resulting success inspired widespread usage of the technique.

However, one advantage of \textbf{TDAsweep} is that to a large extent it obviates the need for data augmentation. For example, consider a situation of vertical flipping. Since the column component counts are the same in a flipped version of an image as in the original, there is no need to add the flipped version to the dataset. The same is true for horizontal flips and row component counts and so on.

Since data augmentation can greatly increase computation time for SVM or whatever other machine learning algorithm is used for classification, this can be a major win for \textbf{TDAsweep}. The more augmented files are added, the longer the algorithm run time, thus the greater the potential advantage of \textbf{TDAsweep}.

(However, in the Tensorflow and Pytorch approach, the software does not actually add new images. Instead, at the batch level, the data are replaced by randomly-transformed versions. This is done at every iteration. Computation time is not appreciably increased under this approach.)

\subsection{Example: Histology Images}
The following shows an example of TDAsweep + liquidSVM on the raw histology-MNIST dataset compared to just the liquidSVM on an augmented histology-MNIST dataset (refer to Table \ref{tda_aug}). For this experiment, the original histology-MNIST dataset consists of 5000 grayscale histology images, each 64 by 64 in size. We obtained the augmented version of histology-MNIST dataset by running data augmentation through flipping, rotation, and shifting, reaching a total sample size of 120,000 histology images, a data expansion factor of 24.

\begin{table}[!ht]
\centering
\caption{{\bf Example of TDAsweep versus image augmentation.} A comparison between the performance of the TDAsweep and image augmentation technique on the Histology Images, each 64x64 in size.}
\begin{tabular}{|l|l|l|l|l|l|l|}
\hline
\multicolumn{1}{|l|}{} & \multicolumn{1}{|l|}{\bf MNIST (raw)} & \multicolumn{1}{|l|}{\bf MNIST (sweep)} & \multicolumn{1}{|l|}{\bf MNIST (sweep)}\\ \hline
Sample size & 5,000 & 120,000 & 5,000  \\ \hline
No. of features & 4096 & 4096 & 382  \\ \hline
intervalWidth & NA & NA & 1  \\ \hline
thresholds & NA & NA & (25,100)  \\ \hline
TDAsweep total time & NA & NA & 169 min  \\ \hline
TDAsweep (core=4) & NA & 12.4 min & 103.76 min  \\ \hline
SVM train time & 1 min & 310.26 min & 0.42 min  \\ \hline
Accuracy (Validation Set) & 57\% & 57.24\% & 70.01\% \\ \hline
\end{tabular}
\label{tda_aug}
\end{table}

Let us examine the results in Table \ref{tda_aug}. Since 5000 is a relatively small sample size, naturally, one would look into data augmentation to increase the data's variation. However, we notice that, even after data augmentation increased the sample size to 120,000, liquidSVM was only able to raise the accuracy by 0.24

As we can see from the results, running \textbf{TDAsweep} on the original dataset with 5000 samples improved the accuracy by a significant amount (roughly 13\% increase), showing how well \textbf{TDAsweep} extracted insightful features from the histology images used in this experiment.

Moreover, \textbf{TDAsweep} + liquidSVM on the original dataset also had a large run-time advantage over liquidSVM on the augmented dataset, 169.42 minutes versus 310.26 minutes. This example demonstrates the strong potential of \textbf{TDAsweep} as a better alternative for smaller datasets requiring data augmentation, both in run-time and accuracy. Note that if our expansion factor had larger than 24, more like the 2048 in AlexNet, presumably the run-time advantage would have been even better.

\section{The "Broken Clock Problem"}
Often a machine learning algorithm, though convergent, will produce the same class prediction for every case, the so-called "broken clock" problem. Remedying such a problem can be challenging.

\subsection{Example: CIFAR-10 Data}
We encountered "broken clock" with the famous CIFAR-10 dataset, which consists of 60,000 color images, each 32x32x3 in size, with ten classes.

\begin{table}[!ht]
\centering
\caption{{\bf Example of TDAsweep against the broken clock problem.} Table demonstrating TDAsweep resolving the SVM broken clock phenomenon on the CIFAR-10 image set}.
\begin{tabular}{|l|l|l|l|l|l|l|}
\hline
\multicolumn{1}{|l|}{} & \multicolumn{1}{|l|}{\bf MNIST (raw)} & \multicolumn{1}{|l|}{\bf MNIST (sweep)} \\\hline

Model Package & liquidSVM & TDAsweep+liquidSVM  \\ \hline
No. of features & 3072 & 1140   \\ \hline
intervalWidth & NA & 1   \\ \hline
thresholds & NA & (25,100)   \\ \hline
Accuracy (Validation Set) & 10\% & 46.86\%  \\ \hline
\end{tabular}
\label{broken_clk}
\end{table}

Since there were 10 classes and the predictions were identical, the accuracy for pure liquidSVM was 10\%. Here \textbf{TDAsweep} succeeded in avoiding the broken clock phenomenon. Of course, one could also avoid such an issue by performing further hyperparameter tuning, and 46.86\% is still not a very good accuracy level for this dataset. Nonetheless, this simple example proves yet another possible advantage of \textbf{TDAsweep}: Dealing with convergence issues.

\section{Other Experiments}
Table \ref{comp_res_noT} shows the results of SVM trained on the various datasets without TDAsweep. Note that the e1071 version was used for Fashion-MNIST, and we used the liquidSVM version for the rest. Table \ref{comp_res_T} shows the results of SVM trained on the various datasets with TDAsweep.

\begin{table}[!ht]
\centering
\caption{{\bf Performance of pure SVM on various image data.} Results of SVM on Fashion-MNIST \cite{xiao2017/online}, kuzushiji-MNIST \cite{clanuwat2018deep}, histology-MNIST (28x28 image version), and EMNIST \cite{cohen_afshar_tapson_schaik_2017}.}
\begin{tabular}{|l|l|l|l|l|l|l|}
\hline
\multicolumn{1}{|l|}{} & \multicolumn{1}{|l|}{\bf MNIST (raw)} & \multicolumn{1}{|l|}{\bf MNIST (sweep)} & \multicolumn{1}{|l|}{\bf MNIST (sweep)}  & \multicolumn{1}{|l|}{\bf MNIST (sweep)}\\ \hline
Model Package & e1071 & liquidSVM & liquidSVM & liquidSVM \\ \hline
Sample size & 70000 & 70000 & 5000 & 118000  \\ \hline
No. of features & 784 & 784 & 784 & 784  \\ \hline
SVM train time & 153 min & 11.45 min & 0.429 min & 105.7 min  \\ \hline
Accuracy (Validation Set) & 89.6\% & 87.59\% & 62.9\% & 82.61\%  \\ \hline
\end{tabular}
\label{comp_res_noT}
\end{table}

\begin{table}[!ht]
\centering
\caption{{\bf Performance of TDAsweep+SVM on vaious image data.} Results of TDAsweep + SVM on Fashion-MNIST, kuzushiji-MNIST, histology-MNIST (28x28 image version), and EMNIST.}.
\begin{tabular}{|l|l|l|l|l|l|l|}
\hline
\multicolumn{1}{|l|}{} & \multicolumn{1}{|l|}{\bf MNIST (raw)} & \multicolumn{1}{|l|}{\bf MNIST (sweep)}\\ \hline
Model Package & liquidSVM & TDAsweep+liquidSVM  \\ \hline
No. of features & 3072 & 1140   \\ \hline
intervalWidth & NA & 1   \\ \hline
thresholds & NA & (25,100)   \\ \hline
Accuracy (Validation Set) & 10\% & 46.86\%  \\ \hline
\end{tabular}
\label{comp_res_T}
\end{table}

Again, we see that TDAsweep has an advantage on overall run-time (TDAsweep + SVM train) if the traditional \textbf{e1071} package is used to train, and we could also observe that TDAsweep could lose that speed advantage when \textbf{liquidSVM} was used. We would like to emphasize that advantage of run-time that TDAsweep brings scales with the complexity of the dataset and the algorithm used to train, as implied by the experiment results. Moreover, even with the smaller datasets and the faster evaluation of training models, our parallelization feature would still accelerate the process and provide the user with other merits such as a superb alternative for data augmentation, a potential cure for the broken clock problem, or simply to avoid overfitting. We believe that the scalability of TDAsweep will be especially usefull in today's research, where larger datasets and more complex models are emerging as the norms. We firmly believe that the multifaceted features TDAsweep have to provide will make TDAsweep a tremendous asset to the Machine Learning research community.


\section{Acknowledgments}
We thank Richard M. Levenson and his lab for providing us useful suggestions and medical datasets for testing; We thank Robin Yancey for her guidance on implementing the parallelization feature for TDAsweep.



\bibliographystyle{unsrt}  


\end{document}